\documentclass[a4paper,twoside]{article}

\usepackage{epsfig}
\usepackage{subfigure}
\usepackage{calc}
\usepackage{amssymb}
\usepackage{amstext}
\usepackage{amsmath}
\usepackage{amsthm}
\usepackage{multicol}
\usepackage{pslatex}
\usepackage{apalike}
\usepackage{SCITEPRESS}     % Please add other packages that you may need BEFORE the SCITEPRESS.sty package.

\usepackage{times}
\usepackage{latexsym}
\usepackage{url}
\usepackage{graphicx}
\usepackage{amsmath}
\usepackage{subfig}

\begin{document}
\title{Exploring the context of recurrent neural network based \\conversational agents}

\author{\authorname{Raffaele Piccini and Gerasimos Spanakis}
\affiliation{Department of Data Science and Knowlednge Engineering, Maastricht University\\Maastricht, 6200MD, Netherlands}
\email{r.piccini@alumni.maastrichtuniversity.nl, jerry.spanakis@maastrichtuniversity.nl}
}


\abstract{
Conversational agents have begun to rise both in the academic (in terms of research) and commercial (in terms of applications) world. This paper investigates the task of building a non-goal driven conversational agent, using neural network generative models and analyzes how the conversation context is handled. It compares a simpler Encoder-Decoder with a Hierarchical Recurrent Encoder-Decoder architecture, which includes an additional module to model the context of the conversation using previous utterances information.
We found that the hierarchical model was able to extract relevant context information and include them in the generation of the output. However, it performed worse (35-40\%) than the simple Encoder-Decoder model regarding both grammatically correct output and meaningful response. Despite these results, experiments demonstrate how conversations about similar topics appear close to each other in the context space due to the increased frequency of specific topic-related words, thus leaving promising directions for future research and how the context of a conversation can be exploited.}

\onecolumn \maketitle \normalsize \vfill
\section{\uppercase{Introduction}}
Interactive conversational agents, also known as Dialogue Systems, or more informally Chatbots, are becoming increasingly popular in recent years among researchers and companies. The reason for this rise in popularity is the wide set of possible applications, ranging from customer services and technical support to video-games and other forms of entertainment.
Additionally, the potential for conversational agents has significantly improved in recent years. Advancements in the field of Machine Learning and the introduction of new Deep Recurrent Neural Networks, made possible the creation of dialogue systems capable of more meaningful interactions with the users such as \cite{serban2016building}.

Conversational agents could be divided (among many other ways) into two categories; goal-driven systems: when the agent has a specific predetermined task to fulfill, such as customer service, and non-goal-driven systems. This paper will deal with the latter category, trying to generate meaningful conversations, with no goal to fulfill, as goal-driven systems would require large corpora of task-specific simulated conversations. Nevertheless, the model might eventually prove useful also for goal-driven tasks.

This paper explores the application of different Encoder-Decoder architectures to generate answers. We compares a simple Encoder-Decoder architecture with a Hierarchical Recurrent Encoder Decoder (HRED) and we seek answers as to how the quality of answers improved. Additionally, the way that HRED network encodes entire conversations in the context latent space will be investigated as to whether it can provide more coherent answers. Finally, an attempt to visualize the context space will be attempted in order to confirm where conversations about similar topics lie. 

\section{\uppercase{Related Work}}
The traditional approach for Conversational Agents follows a modular approach, dividing the process into three modules: a Natural Language Understanding (NLU) unit, a Dialogue Manager and a Natural Language Generation module (NLG). The NLU module will process the input and extract useful information. This information is then used by the Dialogue Manager to update internal states, send a query to a knowledge-based system, or simply follow precoded instructions.
Finally, the NLG will use the information from the Dialogue Manager to generate the output sentence.
The simplest technique used for NLU is to spot certain keywords in the input, often working together with a script-based Dialogue Manager. However, throughout the years there have been many attempts to improve the NLU unit to better extract text information, using techniques including statistical modeling of language \cite{manning1999foundations}, skip-gram models \cite{mikolov2013distributed} and, more recently, deep neural networks \cite{collobert2008unified}.
Eventually, with the rise of Deep Learning in recent years,  Dialogue Systems research has mainly focused on end-to-end models, capable of including all 3 modules in a single deep neural network, trained on a large dataset. One end-to-end RNN architecture proved particularly successful in recent years is the Encoder-Decoder.

The use of Encoder-Decoder architectures for natural language processing was first proposed as a solution for text translation, in 2014 by \cite{cho2014learning}. From then on, the architecture has been applied to many other tasks, including conversational agents \cite{serban2017hierarchical}. However, generating responses was found to be considerably more difficult than translating between languages, probably due to a broader range of possible correct answers to any given input.
A limitation of Encoder-Decoder models to produce meaningful conversations is the fact that any output is only influenced by the latest question. Thus, important factors are ignored, such as the context of the conversation, the speaker, and information provided in previous inputs. In 2015, Sordoni et al. proposed an updated version of the Encoder-Decoder architecture, called Hierarchical Recurrent Encoder Decoder \cite{sordoni2015hierarchical} (HRED), originally used for query suggestions. In their paper, they demonstrate that the architecture is capable of using context information extracted by previous queries to generate more appropriate query suggestions. This paper will attempt to apply such architecture to a dialogue system.

\section{\uppercase{Description of models}}
\subsection{Recurrent Neural Networks}
A Recurrent Neural Network (RNN) is a neural network which works on variable length sequences inputs \textbf{ x } = \textit{(x\textsubscript{1}, x\textsubscript{2} ... x\textsubscript{l})}. Its output consists of a hidden state \textbf{h}, which is updated recursively at each time step by 
\begin{equation}
h(t) = f(h(t-1), x\textsubscript{t})
\end{equation} 
where \textit{f} can be any non-linear activation function. For the scope of this paper \textit{f} was chosen to be a Long Short Term Memory (LSTM) unit.
The reason why RNNs are particularly suited for the task of generating conversations is their ability to receive variable length inputs by encoding them in a fixed length vector. Moreover, they are also able to produce variable length outputs, which is essential when working with sentences which are likely going to be of different lengths. When receiving an input sentence \textbf{ s } = \textit{(w\textsubscript{1}, w\textsubscript{2} ... w\textsubscript{l})} where \textit{l} is the length, the RNN will update its hidden state \textit{h} recursively for every element of the input sequence, in such a way that at any step t, h(t) is a fixed length vector representing all the entire sequence \textit{(w\textsubscript{1}, w\textsubscript{2} ... w\textsubscript{t})} up to time-step $t$. After receiving the last input \textit{$w_l$}, the hidden state will then be representing the entire sequence.

Another interesting property of RNNs is their ability to learn the probability distribution of a sequence. In this specific case, it would learn the probability of a word p to be the next, knowing the previous words in the sentence. 
\begin{equation}
P( w\textsubscript{t} | w\textsubscript{1},w\textsubscript{2}... w\textsubscript{t-1}).
\end{equation}
This can be useful to represent syntactical and grammatical rules and produce believable answers.

\subsection{Word Embeddings}
RNNs cannot accept strings as input, thus a way to encode each word into a numerical vector is required. For this paper, word2vec (using Python's Gensim Library) was used \cite{mikolov2013efficient}. Word2vec uses a neural network to map each word of the vocabulary to a vector space, in which words with similar meaning are closer together. This approach should make it simpler for the chatbot to extract information from the input.
The word2vec model was used with two different approaches.
The first approach was to use a pre-trained word2vec model. This method fixes the embedding matrix, and it does not allow it to be trained with the rest of the model. When using this approach, the output layer uses a linear activation function to output a vector directly in the embedding space. The output vector is then compared to the vectors of the words in the vocabulary, and the closest one is chosen as the output word.
The second approach was to use the pre-trained word2vec model to initialise an embedding layer, connected to the input of the Encoder. The embedding weights would then train with the rest of the network, giving it the possibility to adapt to the specific task. When using this approach, the output layer has the same dimension as the vocabulary, and it uses the softmax activation function to produce a valid probability distribution. The output word is then chosen according to the output probability vector.
The differences in performance between the two approaches are explored in Section \ref{sect:exp}.

\subsection{Encoder-Decoder}
\begin{figure}[h]
\centering
\resizebox{\linewidth}{!}{
\includegraphics{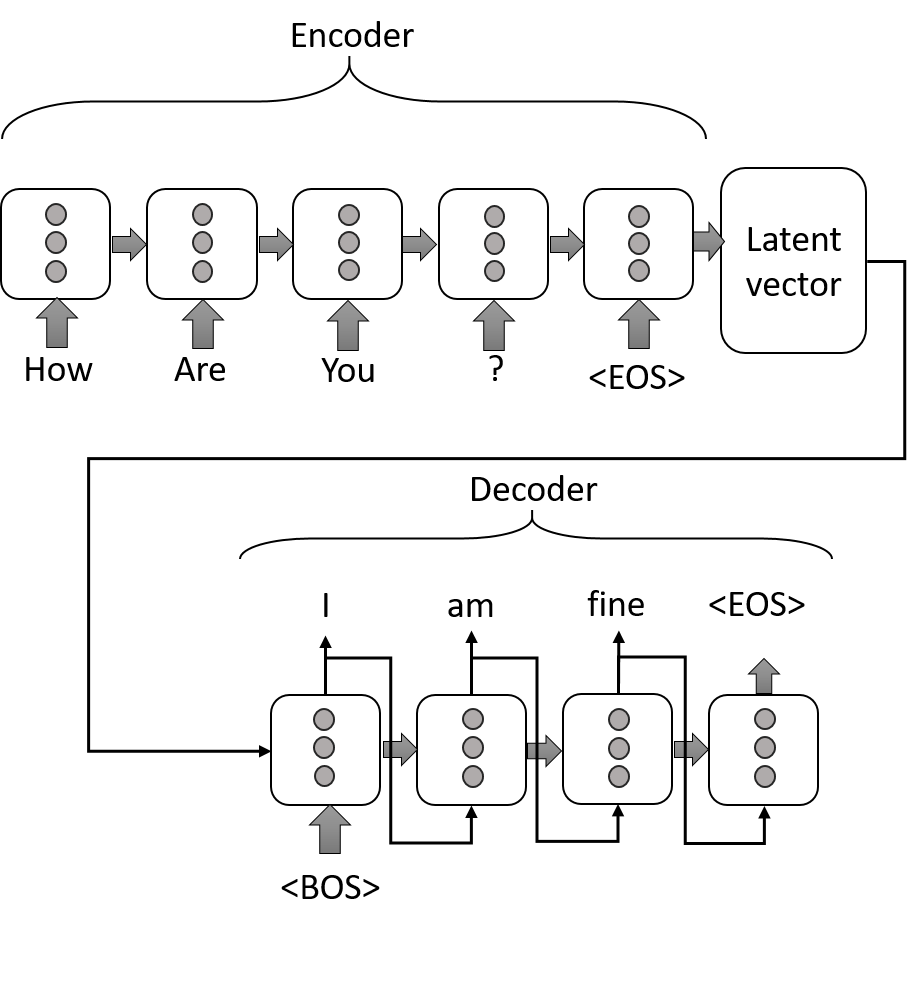}}
\caption{Computational graph of the Encoder-Decoder architecture.}
\label{fig:Enc-Dec}
\end{figure}

An Encoder-Decoder is an architecture (see Figure \ref{fig:Enc-Dec}) that uses two RNNs to learn to encode a variable-length sequence into a fixed-length vector and to decode the fixed-length vector back into a variable-length sequence.

This architecture can be used to learn the conditional probability distribution over a variable-length sequence, conditioned on a different variable-length sequence.
\begin{equation}
P( w1,w2 ... w l\textsubscript{1} | v1,v2... v l\textsubscript{2}).
\end{equation}

The Encoder is a Recurrent Neural Network that reads each word of the input sentence sequentially. After each word, the hidden state of the RNN will be updated according to Eq.(1), until it reads the end-of-sentence token (\textless EOS\textgreater), after which it becomes a vector representation of the entire input sentence. 
The Decoder is a RNN trained to generate the output sentence by predicting the next word in the sentence, conditioned not only by all the previously generated words, but also by the vector representation of the input sentence.
When generating the first word of a sentence, the input will be the begin-of-sequence token (\textless BOS\textgreater). After that, the input will include every word previously predicted by the Decoder. Furthermore, to condition the answer on the input sequence, the RNN is initialised with the hidden state of the encoder when generating each new element.

\subsection{Hierarchical Recurrent Encoder Decoder}
\begin{figure*}[h]
\centering
\scalebox{0.30}{
%\resizebox{\linewidth}{!}{
\includegraphics{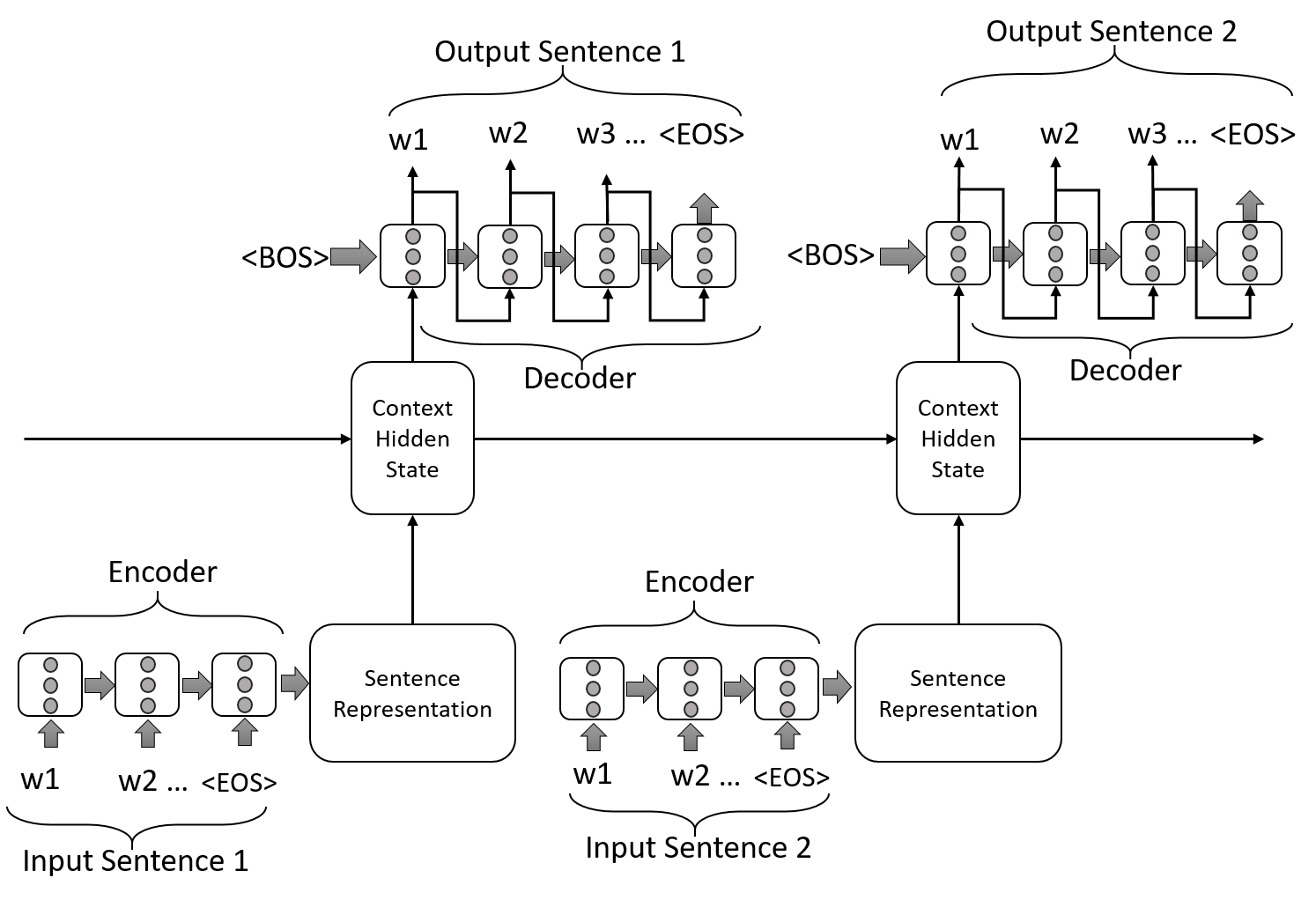}}
\caption{Computational graph of the HRED architecture for a conversation composed of 2 turns.}
\label{fig:HRED}
\end{figure*}

The Hierarchical Recurrent Encoder Decoder (HRED) model is an extension of the simpler Encoder-Decoder architecture (see Figure \ref{fig:HRED}). The HRED attempts to overcome the limitation of the Encoder-Decoder model of generating output based only on the latest input received. The HRED model assumes that the data is structured in a two-level hierarchy: sequences of sub-sequences, and sub-sequences of tokens. In the case of conversational data, each conversation can be modelled as a sequence of sentences, and each sentence as a sequence of words.

The HRED model consists of three different RNNs:  an Encoder, a Decoder, and a context module.
The encoder will handle the data at the lower hierarchical level by processing each word in a sentence, to encode the sentence in a fixed length vector. The sentence vector representation is then given as input to the context module. The context RNN handles the data at the higher hierarchical level by iteratively processing all the sentence vectors of a conversation and updating its hidden state after every sentence. By doing so, the context vector represents the entire conversation up to the last sentence received.
The output of the Context RNN is then used to initialize the hidden state of the Decoder RNN (similarly to the encoder-decoder, where the hidden state of the encoder was used to initialise the decoder), which will generate the output sentence.
The initialization of the Decoder with a vector representing the entire conversation (compared to using the vector of the input sentence) allows HRED to use context information to generate a more appropriate answer. The output is not only conditioned on the input sentence, but also on all the input sentences received during the current conversation session.
HRED is said to be hierarchical as the Encoder and Context RNNs work to encode information at the two hierarchical levels of the conversation. The encoder will try to find short-term correlation within the input sentence, and the context RNN will work at a higher level finding long-term correlations, across different sentences within the same conversation.

\section{\uppercase{Experiments}}
\label{sect:exp}
In this section we describe the dataset used for experimentation followed by the evaluation process of the different models. Finally, all experiments with their results are presented. 

\subsection{Dataset}
In order to test the extraction of context information, a dataset following the hierarchical structure assumed by HRED is needed. The dataset needs to be composed of multiple dialogues, each of which should be a sequence of sentences, as a simple list of questions and answers would not have any context information.
The dataset DailyDialogue is a high-quality multi-turn dialogue dataset, which contains conversations about every-day life \cite{li2017dailydialog} and therefore is ideal for the task of non-goal-driven conversational agents.
Some basic pre-processing was performed on the data-set, although, punctuation was left unchagned, in order for the chat-bot to learn to use it as well.
The DailyDialogue dataset also includes topic labels for each conversation.
These labels can be used to compare the context representations of the conversations. For the experiments, only the 5 most common topics were used. In Section 4.5, experiments will explore whether conversations labelled with the same topics are close together in the context latent space. 

%The code for this paper was written using the python implementation of Keras. Since keras is a very high-level API, it does not support the different hierarchy of input required by the HRED architecture. Notably, it is not possible for the context RNN to process all the sentences vectors of a conversation iteratively, and at the same time for the encoder RNN to process all words of every single sentence of the conversation iteratively. To go around this limitation, after training the network on a sentence, the context state is manually passed to the next time-step, and it is used to initialise the hidden state of the context RNN for the next sentence. The state is then reset to a zero vector at the begin of every new conversation.
%To do so, the function {fit} from Keras API needs to be called on every data-point individually, to allow the context hidden state to be saved and used as input for the next sentence.
%A downside of this training method is that most of the functionalities implemented by keras are not usable. Specifically, increasing the batch size and the number of epochs must be implemented manually.

\subsection{Evaluation Method}
In order to determine the consistency of the output quality, each sentence is evaluated by a user according to two measures:

The first is a binary metric equivalent to whether the output is syntactically correct. This will include the use of punctuation, the use of the end-of-sequence token, and also the correct formulation of a sentence with a valid grammatical structure (including correct use of verbs, adverbs, subjects, names etc.).

The second metric evaluates how appropriate an answer is to the input sentence, and in the case of the hierarchical architecture, also how appropriate it is in the context of the whole conversation. This second evaluation is the most important, especially in comparing the Encoder-Decoder with the HRED architecture. The evaluator will decide whether the response was ``good", ``decent``, or ``wrong".  An output sentence is considered of ``good" quality if it is a valid answer to the input sentence, and if it is pertinent to previous sentences in the conversation.
The quality is considered ``decent" when the answer is a meaningful sentence, and it applies to the input only in part, or not entirely.
A ``wrong" output is when the sentence does not make sense.
Some examples of responses with their evaluations are shown in Table 1.

\begin{table}[htb]
\centering
\caption{Some examples of the evaluation of different outputs for the same input, according to the two metrics described above.}
\resizebox{\linewidth}{!}{
\includegraphics{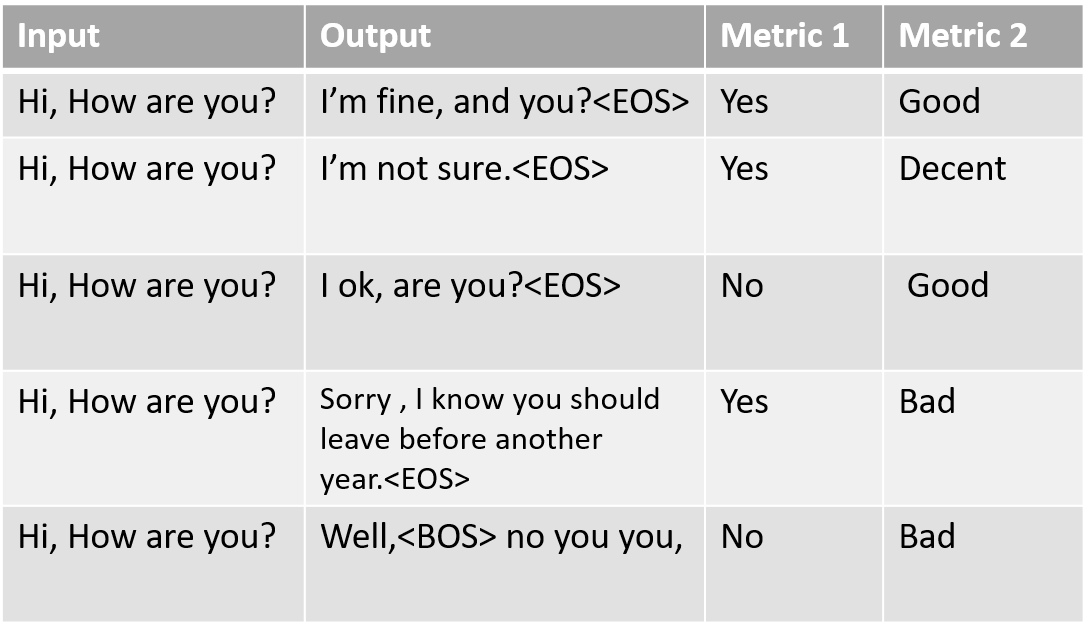}}
\end{table}

Human evaluation was performed by volunteers who were instructed to have a conversation with the bot about a given topic.
A total of 19 volunteers were interviewed, and they performed on average 7 turns of conversation for each one of the 5 topics given.

\subsection{Experimental setup}
The code for this paper was written using the python implementation of Keras. The neural network's parameters are learned by maximising the training log-likelihood using the RMSProp optimiser. For all experiments, the embedding size was tested with a dimension of 100 and 300. Because of GPU memory limitations, the hidden state was fixed to a dimension of 300, and the depth of all the modules was kept to 2.

Initially, the training was performed by trying to predict at each step an entire sentence. This approach resulted in terrible quality outputs, often consisting in the same word repeated or the use of the end-of-sentence token as first output word. It appears that training the model by predicting single words instead gives better results, although this approach will make training significantly slower as the weights are updated after every word and not after every sentence.

Keras require the input of its models to be squared numpy arrays; this requires all conversations in a batch to be of the same length. Similarly, at step i, the i\textsuperscript{th} sentence of all conversations in a batch need to have the same length so that the input vector is a squared matrix.
To do so, the dataset is sorted by the length of their conversation in order to reduce the padding necessary to make all conversations in a batch the same length. In the case of a shorter conversation, a new empty sentence is added (an empty sentence is composed by a begin-of-sentence token immediately followed by an end-of-sentence token). The data also had to be adjusted with zero padding at the level of the individual sentences. Because of memory limitation the batch size was set between 70 to 100.

%\paragraph{Batch training} As previously mentioned, the batch size cannot be increased automatically using keras interface. Alternatively, at each time step, the data are organised in batches manually, thus making the training significantly faster.
%Keras require the input of its models to be squared numpy arrays; this requires all conversations in a batch to be of the same length. Similarly, at step i, the i\textsuperscript{th} sentence of all conversations in a batch need to have the same length so that the input vector is a squared matrix.
%To do so, the dataset is sorted by the length of their conversation in order to reduce the padding necessary to make all conversations in a batch the same length. In the case of a shorter conversation, a new empty sentence is added (an empty sentence is composed by a begin-of-sentence token immediately followed by an end-of-sentence token). The data also had to be adjusted with zero padding at the level of the individual sentences. Because of memory limitation the batch size was set between 70 to 100.

\subsection{Models Comparison}
\paragraph{Embedding}
The goal of the first experiment is to compare the two different ways the word embedding is used to represent the input and outputs.
The first approach consists of using the word2vec model to initialise an embedding layer used as input. The output layer will have the dimension of the vocabulary, and it uses a softmax activation function, to generate a valid probability distribution over every word in the dictionary. The output is then chosen according to the probability vector.
The second approach consists of training a word2vec model separately from the rest of the system, and use it to convert both input and output in n-dimensional vectors (where n is the embedding dimension). The output layer has a linear activation function so that the model predicts vectors in the embedding space. The vector is then compared to all the vectors in the dictionary using cosine similarity to find the closest match.

Both approaches were tested on a simple encoder-decoder architecture and only the best performing version is used for the experiments with the HRED version. The training for both models were performed on a GeForce GTX 960M GPU for about 32 hours. Some examples of responses generated by both models are shown in Table 2. 

\begin{table*}[h!]
\centering 
\caption{Some example responses of the two different approaches. The answer labelled with ``Model 1" is generated by the embedding approach with a probability vector generated by a softmax activation. The one labelled ``Model 2" uses the representation of input and output directly through vectors in the embedding space.}
\resizebox{\linewidth}{!}{
% Table generated by Excel2LaTeX from sheet 'Sheet2'
\begin{tabular}{|p{3cm}|p{21.5em}|}
\hline
\multicolumn{1}{|l|}{\textbf{Input}} & \multicolumn{1}{l|}{\textbf{Output}} \\
\hline
\multicolumn{1}{|r|}{{}{Hey! Do you want to have lunch with me tomorrow after the game?}} & Model 1: I'll meet you, but I need to drive to the ties \textless NUMBER\textgreater, I guess I can, but I won't \\
      & Model 2: You bet, go for it! \\
\hline
\multicolumn{1}{|r|}{{}{Hello, how are you today? It's such a sunny day, I feel like going for a walk}} & Model 1: Perhaps you should leave it to another contract. You never know what \\
      & Model 2: Alright! \\
\hline
\multicolumn{1}{|r|}{{We should meet your brother, I heard he is a funny guy. Is he going to be at the party?}} & Model 1: What about the other? It is more incredibly enjoyable \\
      & Model 2: Well, I'm afraid so. But don't worry about it! \\
\hline
\multicolumn{1}{|r|}{{Hi, how are you?}} & Model 1: I' am sorry and you? \\
      & Model 2: Don't ask me! \\
\hline
\multicolumn{1}{|r|}{{Can you help me find a jacket of my size?}} & Model 1: Well, you know it! Consider it done! \\
      & Model 2: How can I help you! \\
\hline
\end{tabular}%

}
\end{table*}

The results of the experiment show that the second approach learns to generate shorter sentences, which are often extremely vague and mostly grammatically correct. Because of their vagueness, they often apply to the input sentence, thus giving the impression that the answer is somehow appropriate to the input. For this reason, most of the testing volunteers graded the answers generated by this approach higher than the alternative.
On the other hand, the first approach generated sentences that are less often correct from a grammatical point of view, making mistakes including repeating the same word twice when not necessary, and wrong use of punctuation. The response generated by this approach appears to be often a longer and more complex sentence, and although it mostly does have a correct structure, it often mistakenly uses words, making the response less meaningful.

The result of the questionnaire is firmly in favour of the second approach. Nearly all the volunteers (95\%) judged the second chatbot as the best of the two, as it was able to generate a valid answer more often than the first, both grammatically and semantically, thanks to the fact that those answers were usually incredibly vague and could have applied to a substantial number of different inputs. 
Although the second model appears to perform better, a larger dataset and longer training might result in a different conclusion, as neither of the networks seems to have been trained long enough (and on a deep enough network) to learn the semantic relation between words and use them according to their meaning. The second model appears better only because it learned to generate very vague responses.

\begin{table}[htb]
\centering
\caption{Comparison of the accuracy metrics described in subsection 4.2.
}
\resizebox{\linewidth}{!}{
\begin{tabular}{|l|c|c|}
\hline
\textbf{Model} & \textbf{Encoder-Decoder 1} & \textbf{Encoder-Decoder 2} \\
\hline
\hline
Metric1 & 28\%  & 58\% \\
\hline
Metric 2 Good & 12\%  & 30\% \\
\hline
Metric 2 Decent & 20\%  & 31\% \\
\hline
Metric 3 Bad & 67\%  & 38\% \\
\hline
\end{tabular}%
}

\end{table}

\paragraph{Hierarchical model}
The second experiment aims at comparing the Hierarchical Recurrent Encoder-Decoder Architecture, to its simpler version, the Encoder-Decoder model. For this test, both of the networks will use a separately trained word2vec model to convert inputs and outputs into vectors, as it was the method that gave the best results among the two embedding techniques (see Experiment 1). This comparison will reveal whether the Hierarchy structure of the HRED will help it generate more appropriate answers given the context of the conversation.
These models were trained on a GeForce GTX 960M GPU for a total of about 45 hours.

\begin{table*}[h]
\centering
\caption{Some example responses for the encoder-decoder model and HRED model. }
\scalebox{0.75}{
%\resizebox{\linewidth}{!}{
%\includegraphics{Results2}
% Table generated by Excel2LaTeX from sheet 'Sheet3'
\begin{tabular}{|r|r|p{11.165em}|}
\hline
\multicolumn{1}{|p{15.915em}|}{\textbf{Context}} & \multicolumn{1}{p{13.585em}|}{\textbf{Input}} & \textbf{Output} \\
\hline
\multicolumn{1}{|p{15.915em}|}{Did you buy your books for school? Where is the math notebook? $\rightarrow$ I don't want to go to school, I don't like math, it's boring} & \multicolumn{1}{p{13.585em}|}{We have to go, pack your bag} & Enc-Dec: I told you so! \\
      &       & HRED: Please don't worry about it!  \\
\hline
\multicolumn{1}{|p{15.915em}|}{Where are we going for holidays this summer? $\rightarrow$ What about we go in Europe? We can visit museums and go to the beach.} & \multicolumn{1}{p{13.585em}|}{We have to go, pack your bag} & Enc-Dec: I told you so! \\
      &       & HRED: Oh, what do you want to see? I don't there there is anything good German out! \\
\hline
\multicolumn{1}{|p{15.915em}|}{We have a new colleague at work, he is the new operations manager $\rightarrow$ Is he good at his job? $\rightarrow$ No. He doesn't know what he is doing, he was hired because he is friend with the CEO} & \multicolumn{1}{p{13.585em}|}{Are you going to do something about it?} & Enc-Dec: Well, I don't know what to do about it \\
      &       & HRED: I work before Bron. Yankees play tonight. \\
\hline
\multicolumn{1}{|p{15.915em}|}{How is it going with your boyfriend? $\rightarrow$ Six months ago he asked me to marry him and how he cheated on me with another woman, I can't just let it go} & \multicolumn{1}{p{13.585em}|}{Are you going to do something about it?} & Enc-Dec: Well, I don't know what to do about it \\
      &       & HRED: I will get through it \\
\hline
\end{tabular}%

}
\end{table*}

Table 4 displays some examples of outputs generated given an input and a context. The Encoder-Decoder model is the same used in experiment one, the only difference is the amount of training, as for the second experiment the model was trained for a longer time. Although the performance does not seem to change much; The Encoder-Decoder produces very vague answers to any input, so that the output is often a valid response. On the other hand, the HRED tries to produce more specific outputs, also using the high-level information extracted with the context layer. It is clear from the answers generated that the network is able to extract some context information, as there is an increase in the use of words associated with other words in previous sentences. When the context contains a lot of work-related words, the model is more likely to generate outputs with words such as ``work", ``colleague", ``contract", ``office" etc. Similarly, when using words associated with vacations and tourism, some common outputs are names of countries, and words such as ``visiting", ``Museum", ``Hotel" etc...
Unfortunately, although the network is able to extract some context information, it does not seem to be able to use it to form meaningful sentences. Therefore, the result is often a complex response (more complicated than the simplistic and vague answers of the Encoder-Decoder), usually containing quite a lot of grammatical mistakes. In addition, although the words it uses are often appropriate to the context, it fails in using them in a meaningful way, thus generating sentences that often do not make much sense.
Because of these reasons the results of the questionnaire are strongly in favour of the simpler Encoder-Decoder model (88\% of the interviewed people preferred the Encoder-Decoder).

\begin{table}[htb]
\centering
\caption{Comparison of the accuracy metrics described in subsection 4.1.}
\scalebox{0.75}{
%\resizebox{\linewidth}{!}{
\begin{tabular}{|l|c|c|}
\hline
\textbf{Model} & \textbf{HRED} & \textbf{Encoder-Decoder} \\
\hline
\hline
Metric1 & 38\%  & 58\% \\
\hline
Metric 2 Good & 16\%  & 30\% \\
\hline
Metric 2 Decent & 23\%  & 32\% \\
\hline
Metric 3 Bad & 60\%  & 31\% \\
\hline
\end{tabular}%
}
\end{table}

\subsection{Context Representation}
This subsection will include two experiments aiming at investigating the encoding of the sentences in the context space by the context module.
\paragraph{Topics clustering}
The goal of this experiment is to compare vector representations of different conversations (in the context space) in order to determine whether conversations about similar topics are close to each other. After the HRED model is trained, all sentences of the dataset are encoded by the context module in a vector.
Once all conversations are mapped into vectors in the context space, a technique called ``t-distributed stochastic neighbor embedding" \cite{maaten2008visualizing} (t-SNE) is used to visualize the context space in only two dimensions, by finding a 2D representation in which vectors that were close in the original space, are close, and vectors that were distant are further away.

\begin{figure*}[h]
\label{t-SNE}
\centering
\caption{2D visualisation of conversation encodings in the context space. Colours represent conversations about different topics. }
\resizebox{0.9\linewidth}{!}{
\includegraphics{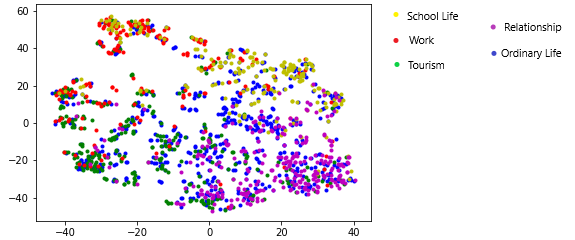}}
\end{figure*}

Figure 3 shows the result of the t-SNE technique to visualise the 100 dimension vectors into 2D.
One can see in the image that, how we would expect, vectors about the same topic are loosely clustered together. This is probably because conversations about the same topics often use the same ``topic-specific" words, thus making the final conversation vectors similar.
The four topic categories ``School", ``Work", ``Relationship" and ``Tourism" are roughly clustered in the four quadrants of the graph, while the "ordinary life" (blue) conversations are much more spread across the entire graph, with a noticeable concentration in the centre-right of the graph. This is likely due to the fact that conversations labelled as ``Ordinary Life" are much more varied, and lack a ``topic-specific" vocabulary if compared to conversation of the other, more specific topics.

\paragraph{Context vector changes during a conversation}
This experiment investigates the way a conversation vector changes in the context space during a conversation, according to the input given.
The average coordinates of the conversation of each topic are calculated,  to better visualise the position of a vector in respect to the clusters of each topic.
A conversation is processed by the model, and after every sentence, the context vector is mapped to the 2D representation of the context space. Doing so one can track the context vector throughout a conversation to see how different inputs affect the position of the context vector in the context space.

\begin{figure*}[h]
\centering
\caption{The black X represents the context vector after every sentence.  The coloured + represent the average coordinate of every topic. }
\resizebox{0.9\linewidth}{!}{
\includegraphics{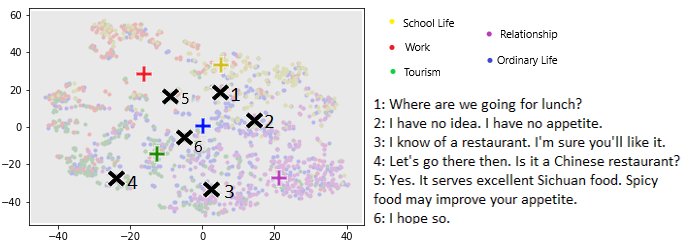}}
\end{figure*}

Figure 4 shows the example of a conversation. One can notice how vague and non-specific sentences tend to push the context vector closer to the centre of the graph, e.g. the last sentence ``I hope so." is a good example of this.
Also in the example, we can notice how naming a nationality, i.e. ``Chinese" in sentence 4, pushes the vector towards the bottom left corner of the graph, close to the average coordinate of the Tourism topic. 
This effect will occur most of the time when the input is a city or a country name.

In order to asses the statistical significance of the aforementioned observations, an additional experiment is performed. The conversation vectors of 150 different conversations, randomly selected from the entire dataset, are mapped in the context vector space. Then, the same input sentence is given to all of the conversations. To establish the effect that the last input had on the context vector, the distance from the centroid of the different topics is calculated both before and after the last input is given. Doing so we can calculate the distance reduction that the last input sentence generated from all of the topic centroids.
Figure 5a illustrates how much the distance of the context vectors of the conversations from all 5 the topic average coordinates, is reduced, after receiving as input the sentence ``I don't know". Because the sentence is general and vague, we expect to see the distance to the ``ordinary life" average coordinate decreasing more compared to the distance from the average coordinates of other topics. 

\begin{figure*}[htb]
\centering
\caption{The Boxplot shows the distance reduction of the conversations vectors from the centroids of the 5 topics, after the input ``I don't know" (in boxplot a) and the tourism related sentence(in boxplot b).}
\resizebox{\linewidth}{!}{
\includegraphics{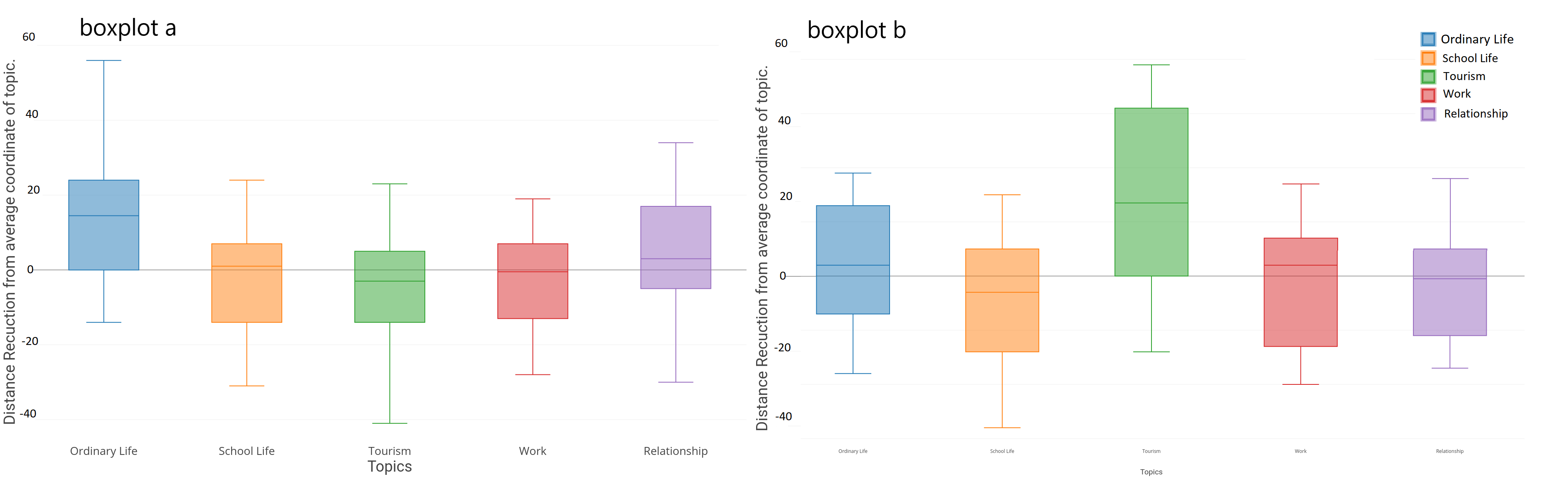}}
\end{figure*}

One can clearly see how in average the context vector moves significantly closer to the ``Ordinary Life'' topic
(moving in average 13.5 units closer), compared to all others (with average ranging from -4.5 to 3).
The same test is performed with the sentence ``Did you do book the flight to China for this summer? I already reserved the hotel". As expected, figure 5b shows a much stronger average decreased distance from the topic ``Tourism" (having a mean distance reduction of 17.1) compared to other topics (with a mean ranging from -4.4 and 3.6), likely because of the intense use of tourism-related words.

In order to ensure that the difference shown in the box-plots are indeed of statistical significance, a Wilcoxon test is performed. The Wilcoxon test was chosen because being a non-parametric test, it does not assume any specific distribution of the data. Table 6 shows the results of the tests. 
\begin{table}[h]
\caption{Table showing the p-value from the Wilcoxon test, on the distance reduction from the 5 topic centroids, after the input of the sentence ``I don't know" in the top triangle, and after the input of the tourism related sentence in the lower triangle.}
\centering
\resizebox{\linewidth }{!}{
\begin{tabular}{c|c|c|c|c|c|}
              & Ordinary Life & School & Work  & Relationship & Tourism \\
              \hline
Ordinary Life & 1             &  \textbf{0.009}  &  \textbf{0.007} &  \textbf{0.03}         &  \textbf{0.001}   \\
School        &0.17       & 1      & 0.39  & 0.18         & 0.63    \\
Work          & 0.71          &0.19    & 1     & 0.44         & 0.27    \\
Relationship  &0.51           &0.26    &0.37   & 1            & 0.09    \\
Tourism   & \textbf{0.01} &\textbf{0.005}&\textbf{0.02}&\textbf{0.009}& 1     
\end{tabular}}
\end{table}

In the top triangle of table 6 it can be noticed that the p-value for all  pairs that include the ``Ordinary Life'' topic are below 0.05, and therefore are significantly different, contrary to all other pairs, which are not significant (the p-value is above 0.05). Similarly lower triangle the p-value for all the pairs involving the ``Tourism'' topic are significant, while all other pairs are not.

The results of this experiment suggest that ``topic-specific" words will strongly influence the position of the context representation in the context latent space.

\section{\uppercase{Conclusion \& Future Work}}
This paper explored the application of a Hierarchical Encoder-Decoder architecture for the task of a Dialogue System. The model attempts to overcome the limitation that the basic Encoder-Decoder model has: the inability of extracting and using context information in the generation of a response. The model that performed the best in the human evaluation experiments is the Encoder-Decoder that uses a separately trained word2vec model to transform the words in vectors.

The reason why Encoder-Decoder outperformed the other models in the experiments, is because it learned to generate simpler and vaguer answers that, due to their simplicity, are more likely to appear as a valid response to the input. The HRED model is still able to extract some contextual information, including the most common words used in all the topics, as demonstrated in experiments 3 and 4. However, it seems that HRED has difficulty learning how to apply such context information to generate meaningful and coherent sentences, resulting in a percentage of grammatically correct sentences generated below 40\%, against almost 60\% for the Encoder-Decoder. 

Even though HRED was not able to improve the quality of the outputs, it was capable of using the context extracted from previous sentences in the generation of the output, resulting in increased usage of words associated with other words previously appeared in the conversation.
HRED has been shown to be able to generate a context space with some level of structure, in which conversations about similar topics are indeed close together. The result of experiment 4 suggests that this effect is probably caused by the fact that conversations labelled with the same topic often use similar words.

Future work will study how a significant increase in the depth of the HRED modules would affect the performance of the model, and what the effect of a much bigger dataset would be on the final result. Additionally, we are planning to explore the application of the architecture to different tasks, as HRED could in principle be applied to any sequence-generating tasks that exhibit a hierarchical structure. In particular, it would be interesting to experiment with hierarchical structures composed by more than 2 levels, by adding one additional context module for every additional level of the hierarchy.

\vfill
\bibliographystyle{apalike}
{\small
\bibliography{bibli}
}
\end{document}